\documentclass[letterpaper]{article} 
\usepackage{aaai2026}  
\usepackage{times}  
\usepackage{helvet}  
\usepackage{courier}  
\usepackage[hyphens]{url}  
\usepackage{graphicx} 
\usepackage{natbib}  
\usepackage{caption} 
\nocopyright
\usepackage{algorithm}
\usepackage{algorithmic}
\usepackage{amsmath}
\usepackage{amssymb}
\usepackage{amsfonts}
\usepackage{dsfont}
\usepackage{booktabs}
\usepackage{multirow}
\usepackage{colortbl}
\usepackage{newfloat}
\usepackage{listings}
\DeclareCaptionStyle{ruled}{labelfont=normalfont,labelsep=colon,strut=off} 
\floatstyle{ruled}
\newfloat{listing}{tb}{lst}{}
\floatname{listing}{Listing}
\title{FERD: \underline{F}airness-\underline{E}nhanced Data-Free \underline{R}obustness \underline{D}istillation}
\author{
    Zhengxiao Li$^1$
    \quad 
    Liming Lu$^1$
    \quad
    Xu Zheng$^{2,3}$
    \quad
    Siyuan Liang$^4$
    \quad
    \\
    Zhenghan Chen$^5$
    \quad
    Yongbin Zhou$^1$
    \quad
    Shuchao Pang$^1$
    }

\affiliations{
    \textsuperscript{\rm 1}Nanjing University of Science and Technology\\
    \textsuperscript{\rm 2}HKUST(GZ)\quad 
    \textsuperscript{\rm 3}INSAIT, Sofia University, St. Kliment Ohridski\\
    \textsuperscript{\rm 4}Nanyang Technological University\\
    \textsuperscript{\rm 5}STCA, Microsoft\\
    
}

\usepackage{bibentry}

\begin{document}

\maketitle

\begin{abstract}
Data-Free Robustness Distillation (DFRD) aims to transfer the robustness from the teacher to the student without accessing the training data. 
While existing methods focus on overall robustness, they overlook the robust fairness issues, leading to severe disparity of robustness across different categories.
In this paper, we find two key problems: (1) student model distilled with equal class proportion data behaves significantly different across distinct categories; and (2) the robustness of student model is not stable across different attacks target.
To bridge these gaps, we present the first Fairness-Enhanced data-free Robustness Distillation (FERD) framework to adjust the proportion and distribution of adversarial examples.
For the proportion, FERD adopts a robustness-guided class reweighting strategy to synthesize more samples for the less robust categories, thereby improving robustness of them.
For the distribution, FERD generates complementary data samples for advanced robustness distillation.
It generates Fairness-Aware Examples (FAEs) by enforcing a uniformity constraint on feature-level predictions, which suppress the dominance of class-specific non-robust features, providing a more balanced representation across all categories. 
Then, FERD constructs Uniform-Target Adversarial Examples (UTAEs) from FAEs by applying a uniform target class constraint to avoid biased attack directions, which distribute the attack targets across all categories and prevents overfitting to specific vulnerable categories.
Extensive experiments on three public datasets show that FERD achieves state-of-the-art worst-class robustness under all adversarial attack (e.g., the worst-class robustness under FGSM and AutoAttack are improved by 15.1\% and 6.4\% using MobileNet-V2 on CIFAR-10), demonstrating superior performance in both robustness and fairness aspects.

\end{abstract}


\section{Introduction}

With the widespread use of Deep Neural Networks (DNNs)~\cite{goswami2018unravelling, gongye2024side, lim2024cutting}, the deployment of lightweight models on edge devices has become increasingly important~\cite{mittal2024comprehensive, min2024lwuavdet, liu2024lightweight}.
However, a large number of studies have shown that these lightweight models are weakly robust in the face of adversarial attacks~\cite{ma2021understanding,li2022review,wei2018transferable,liang2020efficient,liang2022large,liang2022parallel,muxue2023adversarial,wang2023diversifying,liu2023x}.
While traditional adversarial training methods~\cite{jia2022adversarial,Hsiung_2023_CVPR,10478545,10471619,lu2025adversarial,liu2023exploring}, though showing significant advantages on large models, are difficult to achieve desirable results in lightweight models~\cite{wang2024revisiting, ye2019adversarial, huang2021exploring}.
To enhance the defense capability of lightweight models, researchers have proposed the concept of adversarial robustness distillation~\cite{zhang2019theoretically,goldblum2020adversarially,zi2021revisiting,zhu2021reliable,huang2023boosting,yue2024revisiting,zhu2023improving}, which aims to migrate the defense capability of the robust teacher model to the student model, thereby enhancing the latter's performance in an adversarial setting.

However, in practice, raw training data for teacher model are often unavailable, making it difficult to apply traditional distillation methods directly.
To break through this limitation, researchers propose the Data-Free Knowledge Distillation (DFKD) method~\cite{micaelli2019zero,fang2021contrastive,yin2020dreaming,fang2022up}, which synthesizes alternative samples through a training generator to simulate the original data distribution and realize knowledge transfer.
Based on this, the Data-Free Robustness Distillation (DFRD) method~\cite{yuan2024data,wang2024out,zhou2024derd} is further developed to combine the generation and distillation mechanisms to effectively deliver robustness without the need for real data, providing a novel solution for resource-constrained and data-unavailable environments for model defense.

Despite their effectiveness, existing DFRD methods mainly aim to enhance the overall robustness of the student, while overlooking a critical issue: robust fairness~\cite{sun2023improving,yue2023revisiting,zhao2024improving}. 
The robust model may exhibit strong resistance to adversarial attacks on specific categories and remain vulnerable to the others, leading to inconsistent robustness performance across different categories. 
It consequently impacts the reliability and fairness of the model in practical applications.

In this paper, we make the first attempt at investigating the robust fairness problem in the context of DFRD.
We find that although the student tend to inherit the teacher’s class-wise robustness pattern, the inter-class robustness gap is significantly amplified in the distillation process.
We find two phenomena affecting robust fairness: (1) students distilled with equally distributed synthetic data still show class-wise robustness discrepancies; and (2) the success rate of adversarial attacks on students varies significantly depending on the target class.
Based on these findings, we propose a Fairness-Enhanced data-free Robustness Distillation (FERD) framework by adjusting the proportion and distribution of the synthetic samples to mitigate these problems. 

Specifically, for the proportion, we introduce a robustness-guided class reweighting strategy that encourages the generator to synthesize more samples from weakly robust categories, thereby compensating for their vulnerability and promoting fairness in robustness.
For the distribution, we propose to generate complementary data samples for advanced robustness and fairness distillation.
Firstly, we generate Fairness-Aware Examples (FAEs) by enforcing a uniformity constraint on feature-level predictions that are closely associated with non-robust representations. 
This helps suppress the dominance of class-specific non-robust features, ensuring that the benign samples provide a more balanced representation across all classes.
To avoid biased attack directions, we further construct Uniform-Target Adversarial Examples (UTAEs) from FAEs by applying a uniform target class constraint during adversarial generation.
It distributes the attack targets evenly across all categories and prevent overfitting to specific vulnerable category.
We conjecture that robust distillation on adversarial examples with uniformly distributed targets can defend against attacks from different targets.

Experiments on the three datasets (CIFAR-10, CIFAR-100, and Tiny-ImageNet) show that compared with baseline method, FERD improves +15.1\%, +2.7\%, +3.4\% and +6.4\% on the worst-class robustness against FGSM~\cite{goodfellow2014explaining}, PGD-20~\cite{madry2017towards}, $\textup{CW}_{\infty}$~\cite{carlini2017towards}, and AutoAttack (AA)~\cite{croce2020reliable}, respectively, alleviating the robustness bias problem in the DFRD to a certain extent.

Our contributions can be summarized as follows:
(I) We make the first attempt at investigating the problem of robust fairness in DFRD, revealing that uniform distribution among original data categories and attack target bias are the two key factors affecting fairness.
(II) We propose the FERD framework, which enhances robust fairness at both the proportional and distributional levels through robustness-guided category reweighting and distribution-aware sample generation mechanisms.
(III) Experiments have demonstrated that FERD can significantly enhance the robustness and fairness of the student model, and the robust accuracy in the weakest class has been improved by \textbf{+15.1\%} compared with the existing optimal methods, effectively alleviating the robust bias phenomenon in DFRD.

\section{Related Work}
\subsection{Data-Free Robustness Distillation}
DFRD aims to transfer the robustness from teacher to student model using synthetic samples instead of teacher's original training data. 
DFARD~\cite{wang2024out} first defines the concept of DFRD and proves that the difficulty lies in the lower upper bound of knowledge transfer information. They propose an interactive temperature adjustment strategy and an adaptive generator to solve the problem.
DERD~\cite{zhou2024derd} takes a homogenized expert guidance strategy. 
Both clean and robust knowledge are distilled from clean and robust teachers respectively, using the same synthetic data.
To coordinate the gradients of the clean and robust distillation tasks, DERD also introduces a stochastic gradient aggregation module, thereby optimizing the trade-off between robustness and accuracy.
DFHL~\cite{yuan2024data} proposes the concept of High-Entropy Examples (HEEs), which can characterize a more complete shape of the classification boundary.
Distillation on HEEs achieves the best balance between clean accuracy and robustness.
It is worth mentioning that although DFKD doesn't involve robustness, we can transform it into DFRD by adding adversarial noise to synthetic samples during its distillation stage.
For example, Fast~\cite{fang2022up} proposes a fast DFKD, which reuses the common features in the training data to synthesize different data instances.
By generating adversarial examples on synthetic samples, a distilled robust student can be obtained.
In this paper, we make the first attempt at solving the problem of robust fairness transfer in DFRD.

\subsection{Fairness in Robustness}
While enhancing overall model robustness is a common goal, it inevitably leads to significant class-wise performance discrepancies: models become highly robust for some categories while others remain vulnerable to adversarial attacks.
FRL~\cite{xu2021robust} is a pioneering work in highlighting this issue and introduces the concept of ``robust fairness'' to assess such class-wise robustness disparities. 
To address this issue, FRL proposes fairness constraints, adjusting decision margins and sample weights when these constraints are violated.
BAT~\cite{sun2023improving} identify distinct ``source-class'' and ``target-class'' unfairness within adversarial training, tackling these by adjusting per-class attack intensities and applying a uniform distribution constraint. 
Further methods include those by Fair-ARD~\cite{yue2023revisiting}, who improves student model robust fairness by increasing the weights of difficult classes, and ABSLD~\cite{zhao2024improving}, who focuses on adaptively reducing inter-class error risk gaps by modulating the class-wise smoothness of samples' soft labels during training.
In this paper, we introduce FERD to simultaneously enhance model robustness and alleviate robust unfairness problems.

\section{Observation of Fairness in DFRD}
\label{section3}


In this section, we investigate whether robust fairness of the teacher is transferred to the student distilled from DFRD methods. 
We research the robust fairness performance of the models and defense effects against adversarial attacks from different target categories.

\subsection{Robust Fairness of the Student}

\begin{figure}[!t]
\centering
\includegraphics[width=1.0\columnwidth]{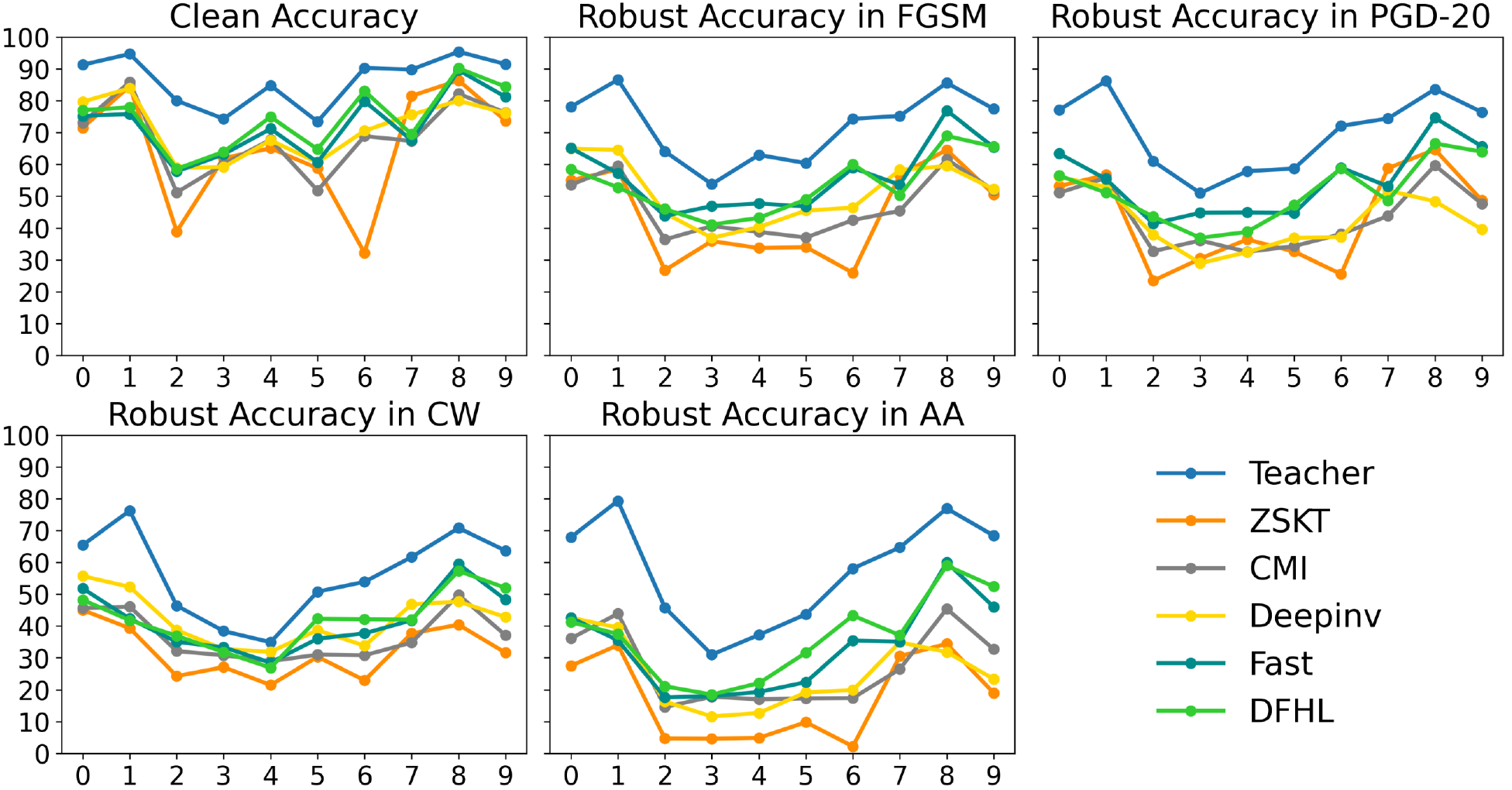} 
\caption{Comparison of the accuracy of the teacher and the students distilled from different DFRD methods under benign and adversarial examples. The blue line represents the teacher and the other lines correspond to the students. The horizontal axis indicates the category, and the vertical axis indicates the accuracy.}
\label{fig1}
\end{figure}

To evaluate the robust fairness of the student, we compare the classification performance of the teacher and the student distilled by five methods (ZSKT~\cite{micaelli2019zero}, CMI~\cite{fang2021contrastive}, DeepInv~\cite{yin2020dreaming}, Fast~\cite{fang2022up}, and DFHL~\cite{yuan2024data}) on different classes. 
We measure the accuracy of each category under benign samples and four adversarial attacks (FGSM, PGD-20, $\textup{CW}_{\infty}$, and AA). The results in Fig.\ref{fig1} show that the performance of the student on different categories follows a consistent trend with that of the teacher. 
Note that the sample labels all adopt a uniform sampling strategy in the aforementioned DFRD methods, in which case the student conducts robust distillation on equally distributed synthetic data. 
However, the robustness among different categories varies significantly. 
Therefore, we argue that sample number imbalance is the key to achieving class balance, because categories behave differently in robust distillation. 
Some categories are more difficult to achieve robustness, while others quickly reach a high level of robustness.
We conjecture that by adjusting the weight of class sampling and increasing the number of samples from weakly robust classes, the problem of robust unfairness can be effectively alleviated.
The proof of this conjecture is shown in Appendix A.

\subsection{Target Fairness of Adversarial Examples}

\begin{figure}[t]
\centering
\includegraphics[scale=0.4]{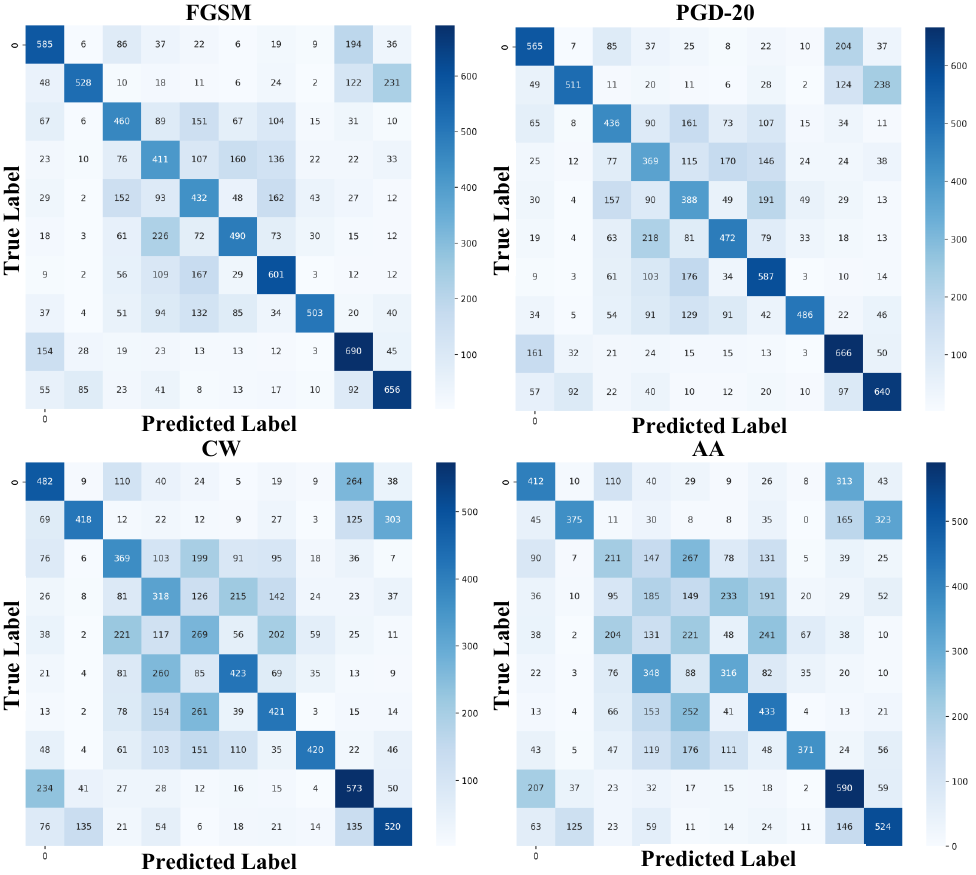}
\caption{Confusion matrices under different adversarial attacks. The horizontal axis denotes predicted labels, and the vertical axis denotes true labels. Darker colors indicate a higher number of samples predicted as the correct class.}
\label{fig2}
\end{figure}

We make confusion matrices to visualize the classification results, as shown in the Fig.\ref{fig2}. 
We observe that the students' robustness against adversarial attacks on different targets varies. 
For samples with the original label of class 0, when the attack target is class 9, the student is more prone to misclassification, showing poor robustness; while for adversarial attacks on other classes, its robustness is relatively strong. 
This indicates that the student exhibits significant differences in defending against adversarial attacks on different targets and are more vulnerable to specific class adversarial attacks.
We conjecture that robust distillation on adversarial examples with uniformly distributed targets can defend against attacks from different targets.
The proof of this conjecture is shown in Appendix B.

\section{Methodology}
\label{section4}

\subsection{Overall Framework}

In the above section, we find two factors affecting robust fairness: (1) the student distilled on data in equal class proportion show class-wise robustness discrepancies; and (2) the success rate of adversarial attacks on the student varies significantly depending on the target class.

\begin{figure*}[h]
\centering
\includegraphics[scale=0.7]{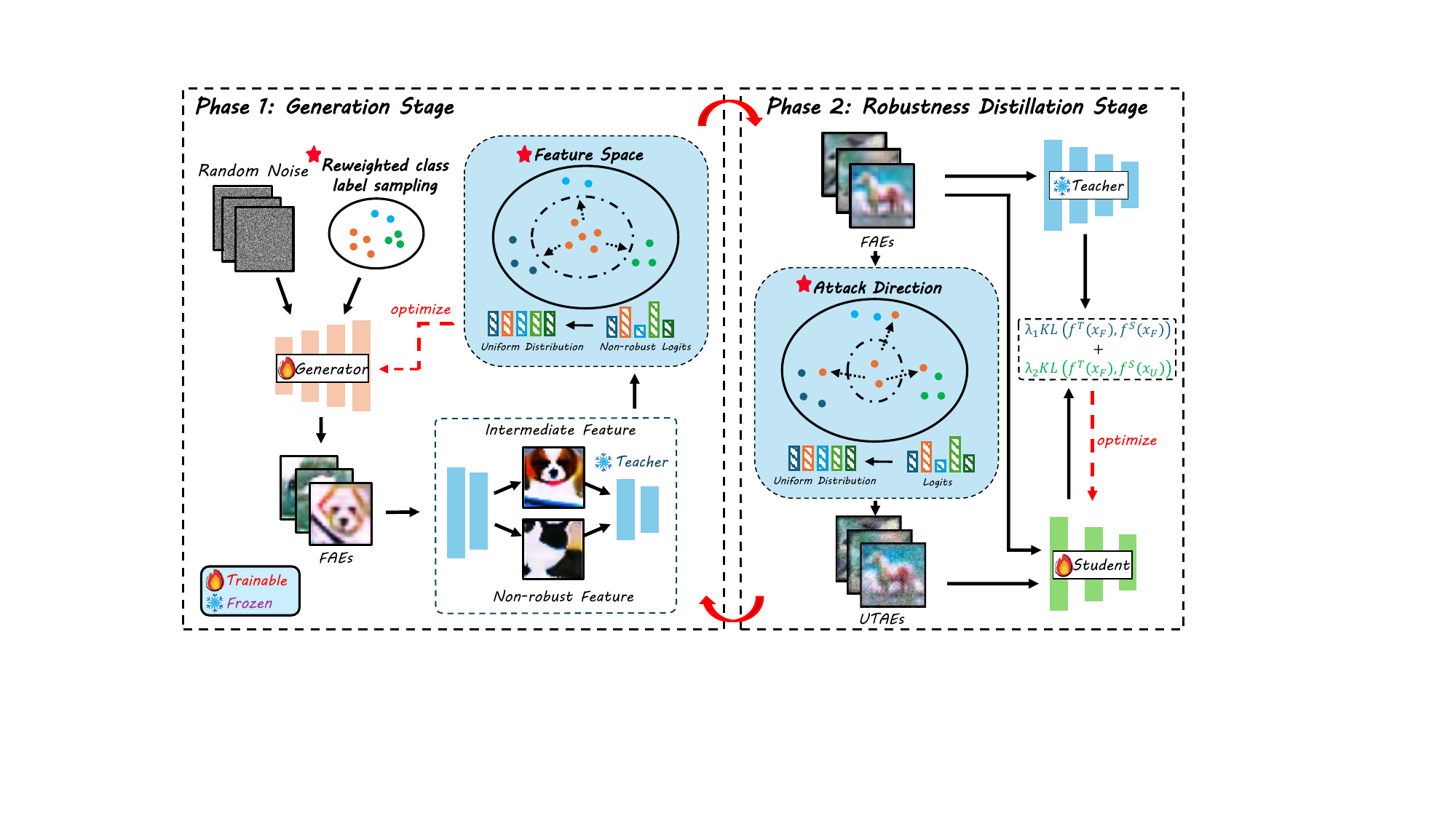}
\caption{Framework of our FERD. In the generation stage, we use a robustness-guided class reweighting strategy to synthesize more weak-class samples and apply a uniformity constraint to non-robust feature predictions to generate FAEs. During robust distillation, we construct UTAEs from FAEs, using them respectively as benign samples and adversarial examples.}
\label{framework}
\end{figure*}

In this section, we introduce our FERD framework and the overall framework is in Fig.\ref{framework}.
The algorithm is shown in the Appendix C.
For the the first proportion problem, we introduce a robustness-guided class reweighting strategy that encourages the generator to synthesize more samples from weakly robust classes. 
For the second distribution problem, we impose uniform constraints on the feature space of synthetic samples and the generation direction of adversarial examples, respectively, making the targets of adversarial examples more uniformly distributed.
Specifically, we firstly generate Fairness-Aware Examples(FAEs) by enforcing constraint on predictions from non-robust feature, which are highly related to adversarial targets.
Then, to avoid biased attack directions, we further construct Uniform-Target Adversarial Examples (UTAEs) from FAEs by applying constraint during adversarial generation, preventing attacks focus on ``vulnerable category''.

\subsection{Class-Aware Sample Reweighting for Fair Distillation}

In traditional DFRD, the labels $y_i$ of synthetic samples $x_i$ are sampled from a uniform distribution, as $y_i \sim \mathcal{U}(0, C-1)$, where $\mathcal{U}$ means uniform distribution and $C$ means class number.
However, robust distillation on equally distributed data infers significant robustness unfairness problem.

To mitigate this problem, we introduce the robustness-guided class reweighting strategy based on adversarial margin, aiming to guide the generator to synthesize more samples of categories with poorer robustness.
Specifically , we first generate adversarial examples $x_i^{adv}$ from the synthetic samples $x_i$ using PGD-20 attack.
Then, we calculate the adversarial margin $m_i$ of each adversarial sample $x_i^{adv}$ under the teacher model $f^T$:
\begin{equation}
m_i = \left( f^T(x_i^{adv}) \right)_{y_i} - \max_{j \ne y_i} \left( f^T(x_i^{adv}) \right)_j.
\end{equation}
The adversarial margin measures the gap between the model's confidence in the correct category and the strongest confusable category. 
The smaller the value, the more probably $x_i^{adv}$ is to be misclassified. 
A negative margin indicates that the attack has been successful.

Therefore, we calculate the average negative adversarial margin for each category:
\begin{equation}
\mathcal{D}c = \frac{1}{N_c} \sum_{i: y_i = c} (-m_i),
\end{equation}
where $N_c$ means the number of samples in category $c$.
This value is used to measure the robustness vulnerability of the category.
A higher value indicates that category $c$ is more likely to be misclassified when facing adversarial attack.

Finally, we apply softmax function to transform $\mathcal{D}c$ into sampling probability distribution $p_c$, which is used as the sampling weight for categories in the subsequent sample generation stage, enabling to adaptively generate more samples of less robust categories.
Furthermore, we compare several other re-weighting strategies. 
The experimental results are shown in Appendix F.

\subsection{Non-Robust Feature Suppression for Balanced Representations}
To achieve the fair tendency to each target when generating adversarial examples, we design a FAEs generation method and use them as benign samples, which ensures that FAEs' non-robust feature predictions are not concentrated in a few categories; otherwise, adversarial perturbations would be more likely to attack such categories. 

We adopt a modified information bottleneck approach to achieve this. The standard information bottleneck objective seeks to learn a compressed representation $Z$ of the input $X$ that is maximally informative about the target label $Y$:
\begin{equation}
\mathcal{L}_{IB} = \mathcal{I}(Z; Y) - \beta \mathcal{I}(Z; X),
\end{equation}
where $\mathcal{I}$ denotes the mutual information and $\beta$ balances the trade-off between prediction accuracy and compression.
We attempt to distill non-robust features $Z_{nr}$ from the intermediate feature representation $Z=f_l(x)$, where $f_l(\cdot)$ describes $l$-th layer outputs of the model. 

When the synthetic samples $x_i$ are input into the teacher $f^T$, we inject a learnable noise scale $\lambda_r$ into $Z$ and define informative features $Z_I$ as follows:
\begin{equation}
Z_I = f^T_l(x_i) + \text{softplus}(\lambda_r) \cdot \epsilon, \quad \epsilon \sim \mathcal{N}(0, I).
\label{eq:non_robust}
\end{equation}
After obtaining $Z_I$ added by $\lambda_r$, we propagate $Z_I$ to the subsequent output layer $f^T_{l+}$ and evaluate the influence of each unit's feature to the teacher prediction. 
When distilling $Z_{nr}$ from $Z_I$, we must ensure $Z_I$ remains predictive while encouraging robustness to noise. 
Since directly optimizing mutual information is intractable, we use a variational approximation~\cite{kim2021distilling}.
The optimization objective is formulated as:
\begin{align}
\min \mathcal{L}(\lambda) =\; & CE(f^T_{l+}(Z_I), y_i) + \beta \cdot \notag \\
& \sum_{c=1}^{Channel} \left( \frac{v_c}{\lambda_c^2} + \log\left( \frac{\lambda_c^2}{v_c} \right) - 1
\right),
\end{align}
where $v_c=\text{Var}(z_r^{(c)})$ denotes the $c$-th channel of $Z_I$ and $\lambda_c$ denotes the $c$-th channel of $\lambda_r$. 
The first term ensures that $Z_I$ correctly predicts the target label $y_i$, while the second term acts as a regularizer to control the amount of information being passed through the bottleneck. 

After optimizing $\lambda_r$, we identify non-robust channels index $i_{nr_k}$ by comparing $\lambda_r^2$ with the maximum variance across all channels:
\begin{equation}
i_{nr_k} = \mathds{1} \left[ \lambda_k^2 < \max_{c' \in \{1, \ldots, C\}} \left\{ \mathrm{Var} \left( Z^{c'}_{I} \right) \right\}
 \right].
\end{equation}
The right-hand side represents the upper limit of perturbation.
Correspondingly, non-robust features $Z_{nr}$ are obtained via channel-wise masking: $Z_{nr} = i_{nr} \cdot Z$. 
The prediction results of $Z_{nr}$ are vulnerable to perturbations, so that they are highly correlated with adversarial predictions. 
To enable the generator to synthesize FAEs, we minimize the $KL$ divergence between the predictions of non-robust features and the uniform distribution:
\begin{equation}
\mathcal{L}_{uni} = KL(\mathcal{U}, f_{l+}^{T}(Z_{nr})).
\end{equation}

To further enhance the quality and diversity of the FAEs, we introduce additional loss functions during the training process of the generator:
\begin{equation}
\left\{
\begin{aligned}
&\mathcal{L}_{adv} = KL\left(f^{T}(x_i), f^{S}(x_i)\right) \\
&\mathcal{L}_{bn} = \sum\nolimits_l \left( \left\| \mu_l(x_i) - \mu_l \right\|_2 + \left\| \sigma_l^2(x_i) - \sigma_l^2 \right\|_2 \right), \\
&\mathcal{L}_{oh} = CE\left(f^{T}(x_i), y_i\right)
\end{aligned}
\right.
\end{equation}
where $\mathcal{L}_{adv}$ encourages divergence between the student and teacher, promoting the diversity of FAEs; $\mathcal{L}_{bn}$ improves the visualization of the FAEs by matching the statistical information (mean $\mu_l$ and variance $\sigma_l^2$) stored in BatchNorm layers of the teacher and the student; $\mathcal{L}_{oh}$ ensures that the FAEs are correctly predicted by the teacher.

Therefore, the overall loss function for the generator in the generation stage are summarized as:
\begin{equation}
\mathcal{L}_{gen} = \lambda_{adv} \cdot \mathcal{L}_{adv} + \lambda_{bn} \cdot \mathcal{L}_{bn} + \lambda_{oh} \cdot \mathcal{L}_{oh} + \lambda_{uni} 
\cdot \mathcal{L}_{uni},
\label{eq:gen}
\end{equation}
where these hyperparameter $\lambda_{adv}$, $\lambda_{bn}$, $\lambda_{oh}$ and $\lambda_{uni}$ are adjusted empirically to balance the trade-offs between robustness, fairness, and accuracy.
The hyper-parameter selection experiments are shown in Appendix G.
By training with the above loss function, the generator synthesize FAEs $x_{F}$, which not only convey effective knowledge, but also be fairer in terms of the tendency towards different categories when generating adversarial examples. 

\subsection{Label-Space Attack Diversification for Fairness Optimization}

To address the problem that the student shows significant differences in defending against adversarial attacks with different targets and is more vulnerable to them from specific categories, we propose a novel adversarial examples generation method, named UTAEs generation.

We aim to construct a more even distributed type of adversarial perturbation, so that the adversarial targets are not limited to certain ``easily misclassified'' classes, but uniformly cover the entire class space. 
To achieve this, we apply a uniform target class constraint during adversarial generation, avoiding to attack from a single direction.
The generation formula is as follows:
\begin{align}
x^{t+1}_{U} &= \Pi_{x_{U} + \mathcal{S}} \Big( x_{U}^t 
+ \alpha \cdot \text{sign} \Big( \nabla_{x^t_U} \big[ KL\left( f^T(x_i), f^T(x^t_{U}) \right) \notag-\\
&\quad \gamma \cdot KL\left( \mathcal{U}, f^T(x^t_{U} \right) \big] \Big) \Big),
\label{eq:utae}
\end{align}
where $\nabla_{x^t_U}$ denotes the gradient of the entropy loss function w.r.t. the UTAE in step $t$ and $\alpha$ is the step size. 
By introducing the $KL$ divergence between $f^T({x_U})$ and $\mathcal{U}$ in the adversarial example generation, we make the target distribution of adversarial examples more extensive.


After synthesizing FAEs ${x_F}$ and UTAEs ${x_U}$, we employ them as benign samples and adversarial examples respectively for robust distillation. 
Here, the robust distillation framework is as follows:
\begin{align}
\mathcal{L}_{stu} =\; & \lambda_1 KL\left(f^{T}(x_F), f^{S}(x_F)\right) \notag + \\
& \lambda_2 KL\left(f^{T}(x_F), f^{S}(x_U)\right).
\label{eq:stu}
\end{align}
Through robust distillation of diversified adversarial targets, the student inherit robustness in a broader range of categories, thereby enhancing overall defensive capability against attacks from all directions.

\section{Experiments}
\label{section:5}
\subsection{Experimental Settings}
\noindent\textbf{Datasets \& Models.} 
We conduct our experiments on three datasets: CIFAR-10~\cite{krizhevsky2009learning}, CIFAR-100, and Tiny-ImageNet~\cite{le2015tiny}. 
For the teacher model, we select WideResNet-34-10~\cite{zagoruyko2016wide} for both CIFAR-10 and CIFAR-100, while PreActResNet-34 is used for Tiny-ImageNet.  
For the student model, we select ResNet-18~\cite{he2016deep} and MoileNet-v2~\cite{sandler2018mobilenetv2} for CIFAR-10 and CIFAR-100. 
For Tiny-ImageNet, we use PreActResNet-18.
The performance of the teacher and the experiments on Tiny-ImageNet are shown in Appendix D and E.

\begin{table*}[t]
  \centering
  \renewcommand{\arraystretch}{1}
  \resizebox{\textwidth}{!}{
  \begin{tabular}{@{}ll ccc ccc ccc ccc ccc!{\vrule width 0pt}@{}}
    \toprule
    \multirow{2}{*}{Student} & \multirow{2}{*}{Method} & \multicolumn{3}{c}{Clean} & \multicolumn{3}{c}{FGSM} & \multicolumn{3}{c}{PGD} & \multicolumn{3}{c}{$\textup{CW}_{\infty}$} & \multicolumn{3}{c}{AA} \\
    \cmidrule(lr){3-5} \cmidrule(lr){6-8} \cmidrule(lr){9-11} \cmidrule(lr){12-14} \cmidrule(lr){15-17}
    & & Avg. & Worst & NSD & Avg. & Worst & NSD & Avg. & Worst & NSD & Avg. & Worst & NSD & Avg. & Worst & NSD \\
    \midrule
    \multirow{6}{*}{RN-18} & ZSKT & 65.48 & 32.20 & 0.280 & 44.17 & 26.00 & 0.324 & 43.06 & 23.50 & 0.349 & 32.02 & 21.50 & 0.256 & 17.16 & 2.20 & 0.780 \\
    & CMI & 68.29 & 51.20 & 0.170 & 46.73 & 36.40 & 0.199 & 43.07 & 32.50 & 0.227 & 36.70 & \underline{28.90} & 0.209 & 26.87 & 14.60 & 0.441 \\
    & DeepInv & 71.22 & \underline{58.90} & \underline{0.130} & 51.34 & 36.90 & 0.196 & 42.22 & 29.00 & 0.221 & 42.11 & 28.10 & \underline{0.198} & 25.19 & 11.60 & 0.445 \\
    & Fast & 72.20 & 57.80 & 0.139 & \underline{56.23} & \underline{43.80} & 0.188 & \underline{54.70} & \underline{37.00} & \underline{0.201} & 41.42 & 28.50 & 0.228 & 33.19 & 17.60 & 0.422 \\
    & DFHL & \underline{74.42} & {58.60} & 0.136 & 53.56 & 41.10 & \underline{0.176} & 51.23 & 36.90 & \textbf{0.198} & \underline{42.15} & 26.90 & 0.214 & \underline{36.37} & \underline{18.50} & \underline{0.369} \\
    & FERD(Ours) & \cellcolor{gray!15}\textbf{79.86} & \cellcolor{gray!15}\textbf{68.20} & \cellcolor{gray!15}\textbf{0.103} & \cellcolor{gray!15}\textbf{61.39} & \cellcolor{gray!15}\textbf{46.90} & \cellcolor{gray!15}\textbf{0.155} & \cellcolor{gray!15}\textbf{55.10} & \cellcolor{gray!15}\textbf{38.60} & \cellcolor{gray!15}\textbf{0.198} & \cellcolor{gray!15}\textbf{46.22} & 
    \cellcolor{gray!15}\textbf{30.50} &
    \cellcolor{gray!15}\textbf{0.191} & 
    \cellcolor{gray!15}\textbf{39.33} &
    \cellcolor{gray!15}\textbf{19.60} &
    \cellcolor{gray!15}\textbf{0.337} \\
    \midrule
    \multirow{6}{*}{MN-V2} & ZSKT & 54.69 & 16.10 & 0.374 & 38.36 & 15.40 & 0.386 & 36.62 & 14.80 & 0.396 & 29.93 & 13.80 & 0.302 & 14.85 & 1.60 & 0.787 \\
    & CMI & 60.70 & 38.60 & 0.216 & 42.37 & 30.00 & 0.227 & 36.81 & 24.70 & 0.261 & 36.39 & \underline{28.50} & 0.209 & 18.08 & 7.60 & 0.532 \\
    & DeepInv & 62.77 & 46.00 & 0.179 & 45.64 & 30.50 & \underline{0.180} & 35.70 & 22.10 & 0.219 & 39.47 & 28.10 & \textbf{0.157} & 15.40 & 5.40 & 0.498 \\
    & Fast & 58.60 & 43.40 & 0.208 & 44.30 & 31.20 & 0.217 & 43.44 & 30.40 & 0.229 & 35.07 & 26.70 & 0.233 & 18.38 & 6.90 & 0.595 \\
    & DFHL & \underline{70.83} & \underline{50.60} & \underline{0.161} & \underline{50.19} & \underline{35.70} & 0.182 & \underline{48.49} & \underline{34.00} & \underline{0.195} & \underline{39.63} & 27.40 & 0.192 & \underline{32.60} & \underline{13.70} & \underline{0.388} \\
    & FERD(Ours) & \cellcolor{gray!15}\textbf{78.04} & \cellcolor{gray!15}\textbf{66.80} & \cellcolor{gray!15}\textbf{0.110} & \cellcolor{gray!15}\textbf{61.44} & \cellcolor{gray!15}\textbf{50.80} & \cellcolor{gray!15}\textbf{0.141} & \cellcolor{gray!15}\textbf{52.55} & \cellcolor{gray!15}\textbf{36.70} & \cellcolor{gray!15}\textbf{0.176} & \cellcolor{gray!15}\textbf{46.21} & \cellcolor{gray!15}\textbf{31.90} & \cellcolor{gray!15}\underline{0.178} & \cellcolor{gray!15}\textbf{38.02} & \cellcolor{gray!15}\textbf{20.10} & \cellcolor{gray!15}\textbf{0.307} \\
    \bottomrule
  \end{tabular}%
  }
  \caption{Result in average robustness(\%) (Avg.~$\uparrow$), worst robustness(\%) (Worst~$\uparrow$), and normalized standard deviation (NSD~$\downarrow$) on CIFAR-10. RN-18 and MN-V2 are abbreviations of ResNet-18 and MobileNet-V2 respectively. The best results are \textbf{bolded}, and the second best results are \underline{underlined}.}
  \label{tab1:cifar10_results}
\end{table*}

\noindent\textbf{Baselines.}
We compare our method with five different DFRD methods, including ZSKT~\cite{micaelli2019zero}, CMI~\cite{fang2021contrastive}, DeepInv~\cite{yin2020dreaming}, Fast~\cite{fang2022up}, and DFHL~\cite{yuan2024data}. 
Note that the first four methods belong to DFKD initially and do not involve robustness. 
We use PGD to generate adversarial examples and then apply the same distillation training loss as RSLAD~\cite{zi2021revisiting} to transform them into DFRD.

\noindent\textbf{Implementation Details.}
Our proposed method and all baselines are implemented on NVIDIA A800 GPU.
The generator is trained via Adam optimizer with a learning rate of 2e-3, $\beta_1$ of 0.5, $\beta_2$ of 0.999. 
The student is trained via SGD optimizer with an initial value of 0.1, momentum of 0.9, and weight decay of 5e-4.
The distillation epochs is set to 220 and the training iterations of generator and student is 400.
The batch size is set to 256 for CIFAR-10 and 512 for both CIFAR-100 and Tiny-ImageNet.
In Eq.\ref{eq:non_robust}, we extract the intermediate features from the last convolutional layer $l$.
In Eq.\ref{eq:gen}, Eq.\ref{eq:utae}, and Eq.\ref{eq:stu}, we set the hyperparameter $\lambda_{adv}$=1, $\lambda_{bn}$=5, $\lambda_{oh}$=1, $\lambda_{uni}$=5,  $\gamma$=0.5, $\lambda_1$=5/6, and $\lambda_2$=1/6.

\begin{figure}[!t]
\centering
\includegraphics[scale=0.3]{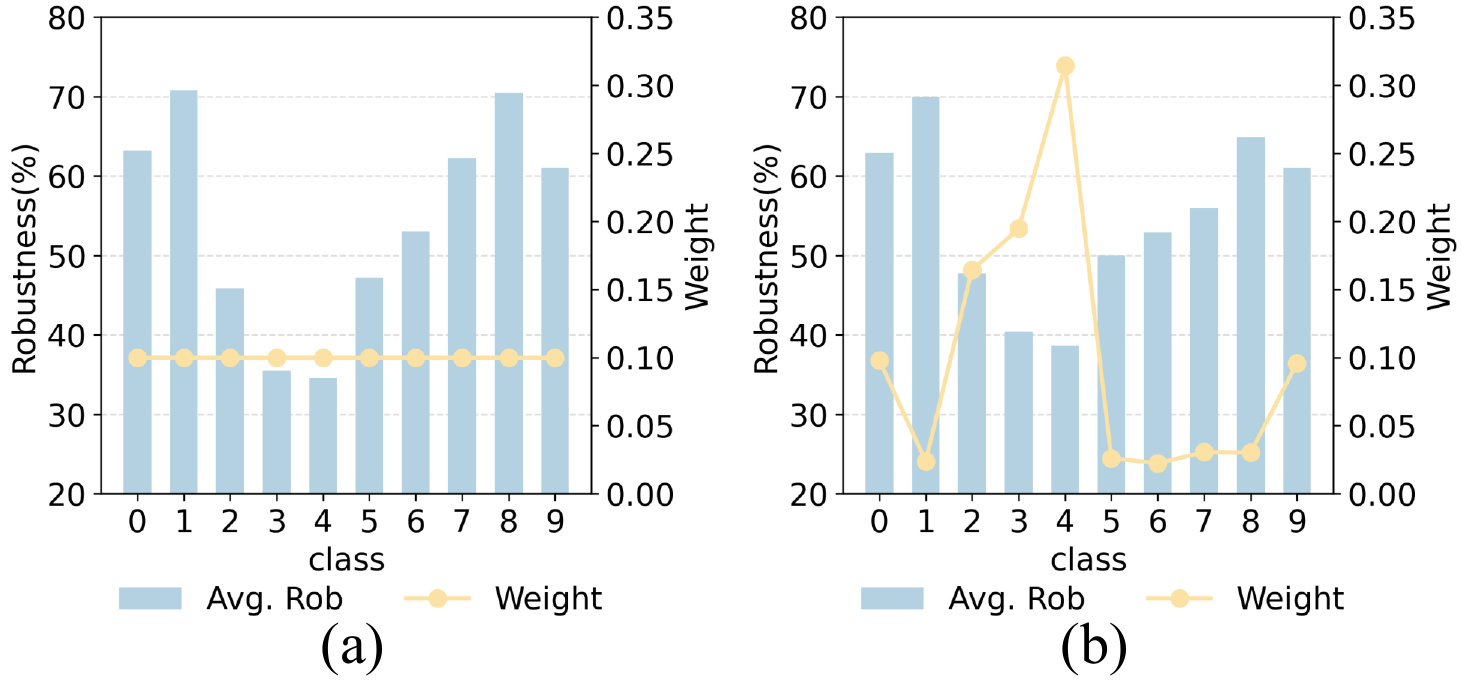}
\caption{(a): The robustness of the student for each category under equal weight. (b): The robustness of the student for each category under re-weighting situation.}
\label{figure_weight}
\end{figure}

\noindent\textbf{Evaluation Metrics.}
We evaluate the robustness of the student against four adversarial attacks: FGSM, PGD, $\textup{CW}_{\infty}$, and AA. 
Follwing~\cite{yue2023revisiting,zhao2024improving}, we employ the worst-class robustness and Normalized Standard Deviation (NSD) to quantify the robust fairness across categories.
NSD is a normalized metric of the adversarial robustness with respect to the standard deviation across different classes. 
The smaller the value of NSD, the better.
Notably, for CIFAR-100 and Tiny-ImageNet, we adopt worst-10\% robustness in place of worst-class robustness, due to the limited size of test set per category and poor performance in the worst class robustness.

\subsection{Experimental Results}

\noindent \textbf{Overall performance.} 
Tab.~\ref{tab1:cifar10_results} and Tab.~\ref{tab2:cifar100_results} show the performance of ResNet-18 and MobileNet-V2 distilled on CIFAR-10 and CIFAR-100 by our method and baselines.
The results demonstrate that the student distilled from FERD has a improvement in the robustness and fairness. 
Our method achieves state-of-the-art worst-class robustness under all attacks. 
When distilled on CIFAR-10 with MobileNet-V2, FERD improves the worst class robustness by 15.1\%, 2.7\%, 3.4\%, and 6.4\% compared with the best baseline against four adversarial attack respectively.
In addition, there is an improvement in the average accuracy and NSD in most of the results. 
Specifically, as shown in Tab.~\ref{tab1:cifar10_results}, FERD achieves the highest average accuracy under all attacks, with an improvement of up to 11.25\%.
Meanwhile, except under the CW$_\infty$ attack, FERD achieves the lowest NSD.

\begin{table*}[t]
  \centering
  \renewcommand{\arraystretch}{1}
  \resizebox{\textwidth}{!}{
  \begin{tabular}{@{}ll ccc ccc ccc ccc ccc!{\vrule width 0pt}@{}}
    \toprule
    \multirow{2}{*}{Student} & \multirow{2}{*}{Method} & \multicolumn{3}{c}{Clean} & \multicolumn{3}{c}{FGSM} & \multicolumn{3}{c}{PGD} & \multicolumn{3}{c}{$\textup{CW}_{\infty}$} & \multicolumn{3}{c}{AA} \\
    \cmidrule(lr){3-5} \cmidrule(lr){6-8} \cmidrule(lr){9-11} \cmidrule(lr){12-14} \cmidrule(lr){15-17}
    & & Avg. & Worst & NSD & Avg. & Worst & NSD & Avg. & Worst & NSD & Avg. & Worst & NSD & Avg. & Worst & NSD \\
    \midrule
    \multirow{6}{*}{RN-18} & ZSKT & 36.44 & 6.40 & 0.522 & 22.15 & 3.50 & 0.573 & 21.70 & 3.20 & 0.582 & 19.08 & 5.10 & 0.491 & 5.98 & 0.00 & 1.362 \\
    & CMI & 51.60 & \underline{27.20} & \underline{0.295} & 35.32 & 14.60 & 0.443 & 32.67 & 12.30 & 0.450 & 27.45 & 10.70 & 0.450 & 17.72 & 1.30 & \underline{0.824} \\
    & DeepInv & \underline{53.44} & \underline{27.20} & 0.298 & 35.51 & \underline{14.90} & \underline{0.422} & 33.76 & \underline{13.40} & \underline{0.436} & 27.73 & 12.10 & \textbf{0.445} & 17.52 & 1.60 & 0.856 \\
    & Fast & 50.69 & 23.50 & 0.323 & \underline{36.11} & 13.10 & 0.440 & \underline{34.97} & 11.80 & 0.443 & 27.73 & 11.20 & 0.456 & \underline{18.27} & \underline{1.70} & 0.859 \\
    & DFHL & 46.27 & 18.70 & 0.370 & 30.54 & 12.70 & 0.449 & 28.31 & 10.90 & 0.485 & \underline{30.54} & 10.60 & \underline{0.449} & 15.23 & 1.10 & 0.957 \\
    & FERD(Ours) & \cellcolor{gray!15}\textbf{56.99} & \cellcolor{gray!15}\textbf{31.00} & \cellcolor{gray!15}\textbf{0.277} & \cellcolor{gray!15}\textbf{39.87} & \cellcolor{gray!15}\textbf{15.60} & \cellcolor{gray!15}\textbf{0.417} & \cellcolor{gray!15}\textbf{37.14} & \cellcolor{gray!15}\textbf{13.60} & \cellcolor{gray!15}\textbf{0.418} & \cellcolor{gray!15}\textbf{30.90} & 
    \cellcolor{gray!15}\textbf{12.40} &
    \cellcolor{gray!15}{0.478} & 
    \cellcolor{gray!15}\textbf{22.33} &
    \cellcolor{gray!15}\textbf{4.40} &
    \cellcolor{gray!15}\textbf{0.783} \\
    \midrule
    \multirow{6}{*}{MN-V2} & ZSKT & 30.77 & 2.70 & 0.650 & 21.38 & 1.50 & 0.680 & 21.00 & 1.40 & 0.688 & 18.86 & 1.90 & 0.612 & 8.67 & 0.00 & 1.204 \\
    & CMI & 35.80 & \underline{13.30} & \underline{0.425} & 23.44 & 6.60 & 0.500 & 20.31 & 5.80 & 0.515 & 20.98 & 7.50 & \underline{0.464} & 8.11 & 0.00 & \underline{1.043} \\
    & DeepInv & 38.10 & 11.10 & 0.447 & 23.42 & 7.20 & 0.555 & 21.80 & 6.30 & 0.568 & 20.58 & 7.50 & 0.524 & 6.67 & 0.00 & 1.420 \\
    & Fast & 39.00 & 13.00 & 0.441 & 26.82 & \underline{8.40} & \underline{0.492} & \underline{25.59} & \underline{7.60} & \underline{0.508} & 22.61 & \underline{9.10} & 0.471 & 9.52 & 0.00 & 1.141 \\
    & DFHL & \underline{42.08} & 11.30 & 0.436 & \underline{27.36} & 8.00 & 0.529 & 25.42 & 6.70 & 0.541 & \underline{23.01} & 1.00 & 0.514 & \underline{12.99} & \underline{0.40} & 1.050 \\
    & FERD(Ours) & \cellcolor{gray!15}\textbf{55.30} & \cellcolor{gray!15}\textbf{29.50} & \cellcolor{gray!15}\textbf{0.286} & \cellcolor{gray!15}\textbf{40.08} & \cellcolor{gray!15}\textbf{16.40} & \cellcolor{gray!15}\textbf{0.397} & \cellcolor{gray!15}\textbf{33.76} & \cellcolor{gray!15}\textbf{11.08} & \cellcolor{gray!15}\textbf{0.463} & \cellcolor{gray!15}\textbf{31.37} & \cellcolor{gray!15}\textbf{12.40} & \cellcolor{gray!15}\textbf{0.462} & \cellcolor{gray!15}\textbf{20.11} & \cellcolor{gray!15}\textbf{1.90} & \cellcolor{gray!15}\textbf{0.810} \\
    \bottomrule
  \end{tabular}%
  } 
  \caption{Result in average robustness(\%) (Avg.~$\uparrow$), worst-10\% robustness(\%) (Worst~$\uparrow$), and normalized standard deviation (NSD~$\downarrow$) on CIFAR-100. RN-18 and MN-V2 are abbreviations of ResNet-18 and MobileNet-V2 respectively. The best results are \textbf{bolded}, and the second best results are \underline{underlined}.}
  \label{tab2:cifar100_results}
\end{table*}

\begin{figure}[!t]
\centering
\includegraphics[scale=0.4]{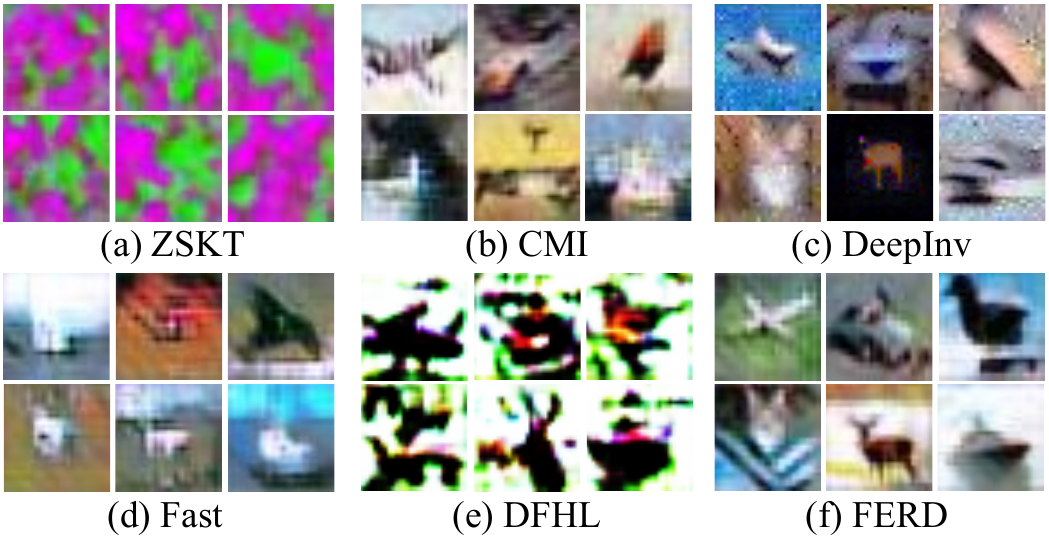}
\caption{32×32 images generated by inverting a WideResNet-34-10 trained on CIFAR-10 with different methods. Clockwise: airplane, car, bird, ship, deer, cat.}
\label{figure_syn}
\end{figure}


\noindent \textbf{Effectiveness of reweighting.} 
Fig.~\ref{figure_weight} illustrates the impact of re-weighted sampling strategy on the robustness of the student model.
The robustness-guided reweighting strategy we proposed effectively identify the categories with poor robustness and increase their sampling weights.
Specifically, the weight of the fourth class which has the lowest robustness reaches 0.314, exceeding other classes.
Correspondingly, the student's robustness in these categories are improved, further enhancing robust fairness.

\noindent \textbf{Synthetic data visualization.} 
In Fig.~\ref{figure_syn}, we visualize the synthetic data generated by FERD and baselines. 
The results indicate that our generator is able to synthesize more visible data. 
In such scenarios, CMI and Fast even suffer from model collapse problem, where the visual quality of the synthetic samples is extremely low. 
This further demonstrates the superiority of our method, which recover high-quality samples from the robust teacher.

\begin{table}[!t]
\centering
\renewcommand{\arraystretch}{1.2}
\begin{tabular}{lcccc}
\toprule
\multirow{2}{*}{Method} & \multicolumn{2}{c}{Clean} & \multicolumn{2}{c}{PGD} \\
\cmidrule(lr){2-3} \cmidrule(lr){4-5}
 & {Avg.} & {Worst} & {Avg.} & {Worst} \\
\hline
FERD & 79.86 & \textbf{68.20} & \textbf{55.10} & \textbf{38.60} \\
w/o reweight & \textbf{80.47} & 63.80 & 54.38 & 34.60 \\
w/o FAEs & 79.06 & 67.30 & 54.03 & 37.50 \\
w/o UTAEs & 78.75 & 65.80 & 54.19 & 37.40 \\
w/o FAEs UTAEs & 78.18 & 62.70 & 53.46 & 36.20 \\
\toprule
\end{tabular}
\caption{Ablation study on components in the framework.}
\label{tab:performance}
\end{table}

\subsection{Ablation Study}

In this section, we provide ablation studies on FERD. 
We keep the same settings with experiments and use WideResNet-34-10 as the teacher, ResNet18 as the student and CIFAR-10 as dataset.

\noindent \textbf{Effectiveness of all components.} 
To verify the effectiveness of our FERD, we conduct ablation studies for each component, and the experimental results are shown in Tab.~\ref{tab:performance}.
We first examine the impact of the re-weighting strategy. 
Compared with the absence of it, the student's accuracy on worst-class significantly improves when re-weighting strategy is applied.
However, it slightly compromise the overall accuracy, which is consistent with the findings of previous researches~\cite{yue2023revisiting, zhao2024improving}.
Further, we analyze the contributions of FAEs and UTAEs. 
Removing either component individually results in a consistent performance decline across all metrics.
When both FAEs and UTAEs are removed simultaneously, the performance further deteriorates. 
This confirms that FAEs and UTAEs are complementary and their joint effect is crucial for achieving the best robust fairness performance.

\begin{figure}[!t]
\centering
\includegraphics[scale=0.6]{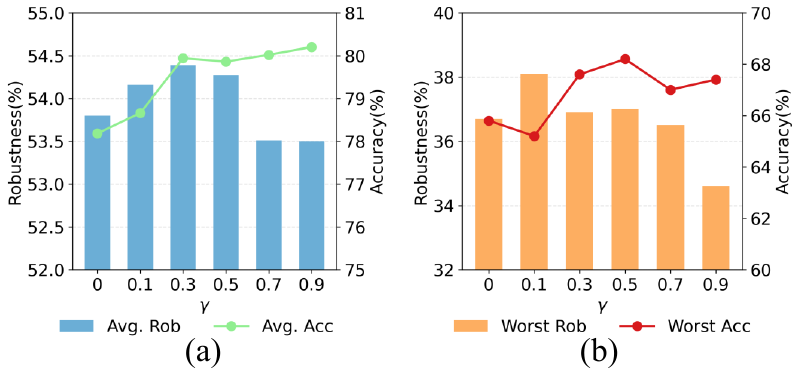}
\caption{Ablation study. (a): The average robustness and accuracy of the student under different $\gamma$. (b):The worst-class robustness and accuracy of the student under different $\gamma$.}
\label{figure5}
\end{figure}

\noindent \textbf{Hyer-param $\gamma$.} 
In Fig.\ref{figure5} (a) and (b), we illustrate the average and worst-class performance of the student distilled with with varying $\gamma$ during UTAEs generation. 
$\gamma$ controls the strength of the uniform target class constraint during adversarial examples generation.
Note that when $\gamma$=0, the adversarial examples are the same as the standard PGD method.  
For average robustness and average accuracy, we observe that a low to medium $\gamma$ (e.g., 0.1 to 0.5) positively enhances the robustness and fairness of the student. 
This indicates the uniform constraint enhances the robustness of the model without significantly affecting the intensity of adversarial perturbations.
However, A high $\gamma$ (e.g., 0.7 or 0.9) enforces strong uniform constraint, which overly suppress the optimization of the adversarial loss, leading to weaker perturbations that reduce attack strength of the adversarial examples.
Consequently, this negatively affects the robust distillation process and leads to a reduction in overall robustness.

\section{Conclusion}
In this paper, we made the first attempt to investigate the robust fairness in DFRD. 
We summarized two key factors affecting robust fairness and propose a FERD framework to mitigate these problems by adjusting the proportion and distribution of adversarial examples.
For the proportion, we introduced a robustness-guided class reweighting strategy to encourage the generator to synthesize more samples from weakly robust classes.
For the distribution, we designed FAEs and UTAEs, taking them as benign samples and adversarial examples respectively for robust distillation.
Extensive experiments show that FERD significantly improves the robust and fairness performance of the student model.
Our work is more applicable and can be effectively applied in practical scenarios.

\bibliography{aaai2026}
\end{document}